\documentclass[runningheads]{llncs}
\usepackage{graphicx}
\usepackage{multirow}
\usepackage{tikz}
\usepackage{comment}
\usepackage{amsmath,amssymb} 
\usepackage{color}
\usepackage{float}
\usepackage{marvosym}
\usepackage{paralist}
\usepackage{enumitem}
\usepackage{colortbl}
\usepackage{booktabs}
\usepackage{subcaption}
\newcommand{\etal}{\textit{et al.}}
\newcommand{\mpage}[2]
{
\begin{minipage}{#1\linewidth}\centering
#2
\end{minipage}
}

\usepackage[pagebackref=true,breaklinks=true,colorlinks,bookmarks=false]{hyperref}

\usepackage[accsupp]{axessibility}  

\begin{document}
\pagestyle{headings}
\mainmatter
\def\ECCVSubNumber{1456}  

\title{Flow-Guided Transformer for Video Inpainting} 

\titlerunning{Flow-Guided Transformer for Video Inpainting}
%
\author{Kaidong Zhang\inst{1} \and
Jingjing Fu \inst{2}\textsuperscript{(\Letter)} \and
Dong Liu \inst{1}}
\authorrunning{K. Zhang et al.}
%
\institute{$^1$University of Science and Technology of China~\quad~$^2$Microsoft Research Asia
\email{richu@mail.ustc.edu.cn, jifu@microsoft.com, dongeliu@ustc.edu.cn}}

\maketitle
\begin{figure}[ht]
\hsize=\textwidth
\centering
\includegraphics[width=0.95\linewidth]{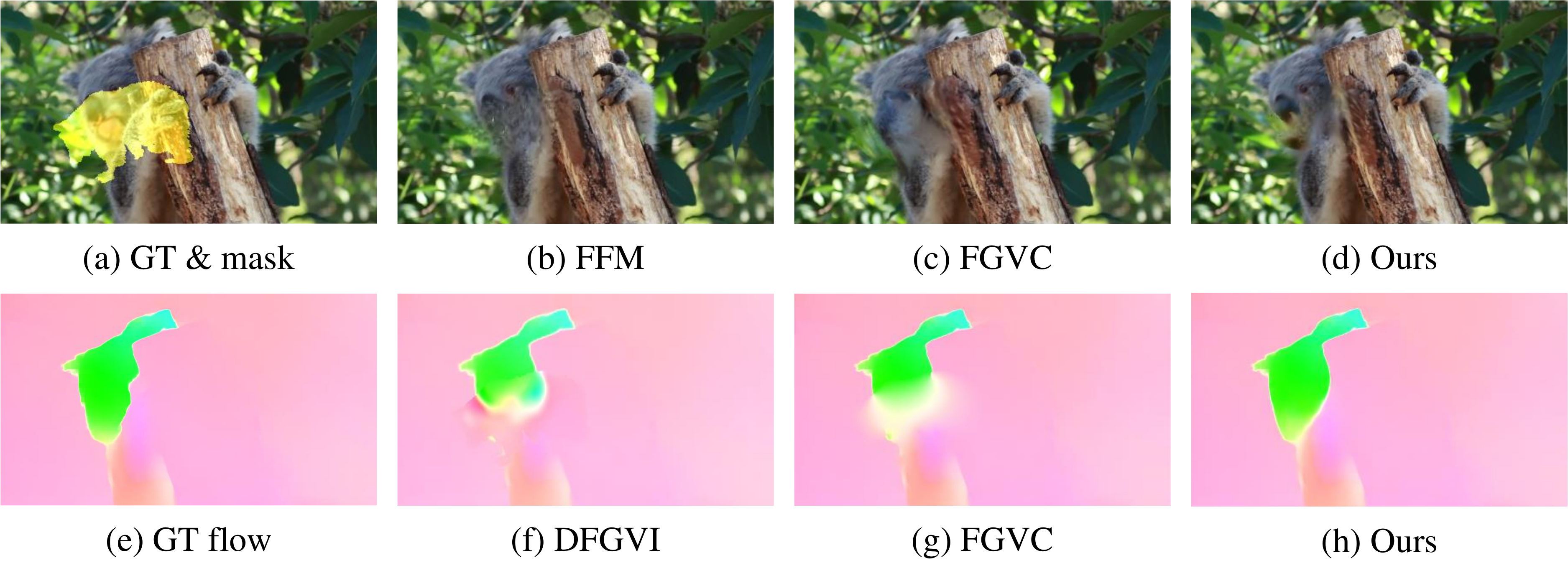}
\caption{Performance comparison on frame synthesis (top row) and flow completion (bottom row) between our method and some state-of-the-art baselines \cite{Xu_2019_CVPR,Gao-ECCV-FGVC,Liu_2021_FuseFormer}. Our method achieves significant performance improvement against the compared baselines and obtains more coherent results.}
\label{fig:teaser}
\end{figure}



\begin{abstract}
We propose a flow-guided transformer, which innovatively leverage the motion discrepancy exposed by optical flows to instruct the attention retrieval in transformer for high fidelity video inpainting. More specially, we design a novel flow completion network to complete the corrupted flows by exploiting the relevant flow features in a local temporal window. With the completed flows, we propagate the content across video frames, and adopt the flow-guided transformer to synthesize the rest corrupted regions. We decouple transformers along temporal and spatial dimension, so that we can easily integrate the locally relevant completed flows to instruct spatial attention only. Furthermore, we design a flow-reweight module to precisely control the impact of completed flows on each spatial transformer. For the sake of efficiency, we introduce window partition strategy to both spatial and temporal transformers. Especially in spatial transformer, we design a dual perspective spatial MHSA, which integrates the global tokens to the window-based attention. Extensive experiments demonstrate the effectiveness of the proposed method qualitatively and quantitatively. Codes are available at \href{https://github.com/hitachinsk/FGT}{{https://github.com/hitachinsk/FGT}}.
\keywords{Video inpainting, Optical flow, Transformer}  
\end{abstract}

\section{Introduction}
Video inpainting aims at filling the corrupted regions in a video with reasonable and spatiotemporally coherent content \cite{bertalmio2001navier}. Its application includes but not limited to watermark removal \cite{newson2014video}, object removal \cite{10.1111/j.1467-8659.2012.03000.x}, video retargeting \cite{kim2019deep}, and video stabilization \cite{1634345}. High-quality video inpainting is challenging because it requires spatiotemporal consistency of the restored video. Directly applying image inpainting methods \cite{pathakCVPR16context,IizukaSIGGRAPH2017,liu2018partialinpainting,yu2018generative,yu2018free,Nazeri_2019_ICCV,9113276,peng2021generating,Liao_2021_CVPR} is sub-optimal, because they mainly refer the content within one frame but fail to utilize the complementary content across the whole video.

Transformer \cite{vaswani2017attention} has sparked the computer vision community. Its outstanding long-range modeling capacity makes it naturally suitable for video inpainting, as video inpainting relies on the content propagation across frames spatiotemporally to fill the corrupted regions with high fidelity. Previous works \cite{yan2020sttn,Liu_2021_FuseFormer,liu2021decoupled} modify transformer for video inpainting task, and achieve unprecedented performance. However, these works still suffer from inaccurate attention retrieval. They mainly utilize the appearance features in transformer, but ignore the object integrity exposed by the motion fields, which indicates the relevant regions.

Recently, several works \cite{Xu_2019_CVPR,Gao-ECCV-FGVC,Zhang_2022_CVPR} propose to complete optical flows for video inpainting. As discussed in DFGVI \cite{Xu_2019_CVPR}, optical flows are much easier to complete because they contain far less complex patterns than frames. Since the relative motion magnitude between foreground objects and background are different, the contents with similar motion pattern are more likely to be relevant. Therefore, the motion discrepancy of optical flows can serve as a strong instructor to guide the attention retrieval for more relevant content. Inspired by this, we propose a novel flow-guided transformer to synthesize the corrupted regions with the motion guidance from completed flows. Our method contains two parts: the first is a flow completion network designed to complete the corrupted flows, and the second is the flow-guided transformer proposed to synthesize the corrupted frames under the guidance of the completed flows.

During flow completion, we observe that the flows in a local temporal window are more correlated than the distant ones, because motion fields are likely to be maintained in a short temporal window. Therefore, we propose to exploit the correlation of complementary features of optical flows in a local temporal window, which is different from the simply stacking strategy in DFGVI \cite{Xu_2019_CVPR} and the single flow completion method in FGVC \cite{Gao-ECCV-FGVC}. We integrate spatial-temporal decoupled P3D blocks \cite{qiu2017learning} to a simple U-Net \cite{ronneberger2015u}, which completes the target flow based on the local reference flows. Furthermore, we propose a novel edge loss to supervise the completion quality in the edge regions without introducing additional computation cost during inference. Compared with previous counterparts \cite{Xu_2019_CVPR,Gao-ECCV-FGVC}, our method can complete more accurate flows.

Under guidance of the completed optical flows, we propagate the content from the valid regions to the corrupted regions, and then synthesize the rest corrupted content in the video frames with the flow-guided transformer. Following previous transformer-based video inpainting methods \cite{yan2020sttn,Liu_2021_FuseFormer,liu2021decoupled}, we sample video frames from the whole video and inpaint these frames simultaneously. Given that the optical flows are locally correlated, we decouple the spatial and temporal attention in transformer and only integrate optical flows into spatial transformers. In temporal transformer, we perform multi-head self-attention (MHSA) spatiotemporally, while in spatial transformer, we only perform MHSA within the tokens coming from the same frame. Considering that the completed flows are not perfect and the content with different appearance may have similar motion patterns, we propose a novel flow-reweight module to control the impact of flow tokens based on the interaction between frame and flow tokens adaptively.

To improve the efficiency of our transformer, we introduce window partition strategy \cite{Liu_2021_ICCV,yang2021focal,chu2021Twins} in the flow-guided transformer. In temporal transformer, as the temporal offset between distant frames may be large, small temporal window size cannot include abundant temporal relevant tokens. As a result, we perform MHSA in a large window to exploit rich spatiotemporal tokens. In spatial transformer, after flow guidance integration, we restrict the attention within a smaller window based on local smoothness prior of natural images. However, simple window attention ignores the possible correlated content at the distant location. To relieve such problem, we extract the tokens from the whole token map globally and integrate these global tokens to the key and value. In such manner, the queries can not only retrieve the fine-grained local tokens, but also attend to the global content. We refer this design as dual perspective spatial MHSA.

 We conduct extensive experiments to validate the effectiveness of different components of our method. As shown in Fig.~\ref{fig:teaser}, our method remarkably outperforms previous baselines in terms of visualization results on frame synthesis and flow completion. In summary, our contributions are:
\begin{itemize}
    \item We propose a flow-guided transformer to integrate the completed optical flow into the transformer for more accurate attention retrieval in video inpainting.
    \item We design a novel flow completion network with local flow features exploitation, which outperforms previous methods significantly.
    \item We introduce window partition strategy in the video inpainting transformer and propose the dual perspective spatial MHSA to enrich the local window attention with global content.
\end{itemize}

\section{Related Work}
\textbf{Traditional methods.} Traditional video inpainting methods \cite{bertalmio2001navier,10.1111/j.1467-8659.2012.03000.x,1634345,7112116,granados2012b,newson2014video} explore the geometry relationship (e.g. homography or optical flows) between the corrupted regions of the target frames and the valid regions of the reference frames for content synthesis with high fidelity. Huang \etal \cite{Huang-SigAsia-2016} design a set of energy equation to optimize optical flow reconstruction and frame synthesis interactively and achieve unprecedented video inpainting quality.

\noindent\textbf{Deep learning based methods.} Deep learning based methods can be divided into two categories, the first one \cite{Xu_2019_CVPR,Gao-ECCV-FGVC} aims to complete the missing optical flows to capture the motion correlation between the valid regions and the corrupted regions. Our method also includes the flow completion component, but we only exploit the complementary flow features in a local window for more efficient and accurate flow completion.

The second category targets on directly synthesizing the corrupted regions from video frames. Some works adopt 3D CNN \cite{wang2019video,chang2019free}  or channel shift \cite{chang2019learnable,zou2020progressive,ke2021voin} to model the complementary features between local frames. Several methods integrate recurrent \cite{kim2019deep,Li_2020} or attention \cite{lee2019cpnet,Oh_2019_ICCV} mechanism into CNN-based networks to expand the temporal receptive field. Inspired by the spatiotemporal redundancy in videos, Zhang \etal \cite{zhang2019internal} and Ouyang \etal \cite{ouyang2021video} adopt internal learning to perform long range propagation for video inpainting. Currently, Zeng \etal \cite{yan2020sttn} and Liu \etal \cite{Liu_2021_FuseFormer,liu2021decoupled} design specific transformer \cite{vaswani2017attention}  to retrieve similar features in a considerable temporal receptive field for high-quality video inpainting. Our method is also built upon transformer, but differently we improve the attention retrieval accuracy with the completed flows.

\noindent\textbf{Transformer in vision.} Due to the outstanding long range feature capture ability, transformer \cite{vaswani2017attention} has been introduced to various computer vision tasks, such as basic architecture design \cite{Liu_2021_ICCV,yang2021focal,chu2021Twins}, image classification \cite{dosovitskiy2021an,Bhojanapalli_2021_ICCV,Fan_2021_ICCV,Wu_2021_ICCV}, object detection \cite{carion2020end,Misra_2021_ICCV}, action detection \cite{Wang_2021_ICCV}, segmentation \cite{wang2021max}, etc. We revisit the design of transformer in video inpainting and propose 
several strategies to improve efficiency while maintaining competitive performance, including spatial-temporal decomposition and the combination of local and global tokens.

\begin{figure*}[t]
\begin{center}
\includegraphics[width=\linewidth]{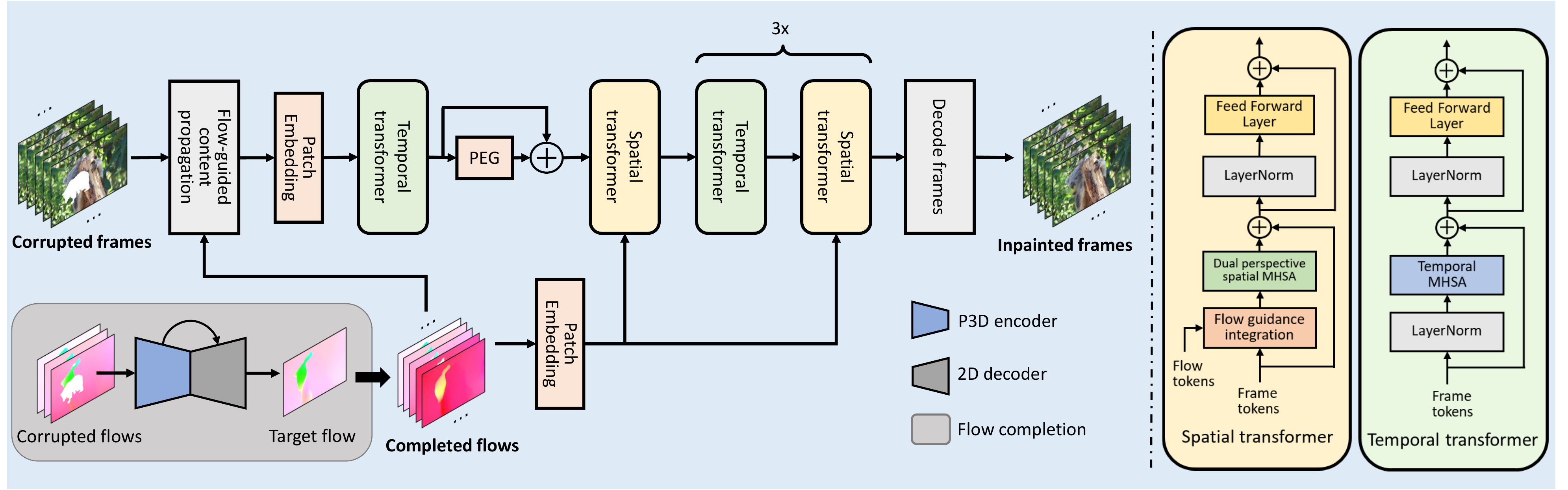}
\end{center}
   \caption{Our method consists of two steps. Firstly, we adopt the \textbf{L}ocal \textbf{A}ggregation \textbf{F}low \textbf{C}ompletion network (LAFC) to complete the corrupted target flow, and then propagate the content among the video frames with the completed flows. Secondly, we synthesize the rest corrupted regions with \textbf{F}low-\textbf{G}uided \textbf{T}ransformer (FGT). PEG: Position embedding generator.}
\label{fig:pipeline}
\end{figure*}

\section{Method}
\subsection{Problem formulation}
Given a corrupted video sequence $X:=$\{$X_1, ..., X_T$\}, whose corrupted regions are annotated by the corresponding mask sequence $M:=$\{$M_1,...,M_T$\}. $T$ is video length. Our goal is to generate the inpainted video sequence $\hat{Y}:=$\{$\hat{Y}_1, ..., \hat{Y}_T$\} and maintain the spatiotemporal coherence between our result and the ground truth video sequence $Y:=$\{$Y_1,..., Y_T$\}.

\subsection{Network overview}
As shown in Fig.~\ref{fig:pipeline}, our network consists of a \textbf{L}ocal \textbf{A}ggregation \textbf{F}low \textbf{C}ompletion network (LAFC) for flow completion and a \textbf{F}low-\textbf{G}uided \textbf{T}ransformer (FGT) to synthesize the corrupted regions. For a given masked video sequence $X$, we extract its forward and backward optical flows $\Tilde{F}_f$ and $\Tilde{F}_b$ and utilize LAFC to complete each optical flow with its local references. Based on completed flows, we propagate the content across video frames. As for the rest corrupted regions, we adopt FGT to synthesize them.

\subsection{Local aggregation flow completion network} \label{p3d-unet}
\noindent\textbf{Local flow aggregation.} The motion direction and velocity of objects vary overtime, and the correlation between distant optical flows may be degraded severely. Fortunately, the variance of motion in short time is a gradual process, which means optical flows in a short temporal window are highly correlated, and they are reliable references for more accurate flow completion.

3D convolution block \cite{tran2015learning} is suitable to capture the local relevant content spatiotemporally. However, the parameter and computation overhead of 3D convolution block are large, which increases the difficulty for network optimization. Considering efficiency, we adopt P3D block \cite{qiu2017learning} instead to decouple the local flow feature aggregation along temporal and spatial dimension. We insert P3D blocks to the encoder of LAFC and add skip connection \cite{ronneberger2015u} to exploit the local correlation between flows. Considering that LAFC completes forward and backward optical flows in the same manner, we denote both $F_f$ and $F_b$ as $F$ for simplicity. Given a corrupted flow sequence, we utilize Laplacian filling to obtain the initialized flows $\Tilde{F}$=\{$\Tilde{F}_{t-ni}, ..., \Tilde{F}_{t}, ..., \Tilde{F}_{t+ni}$\}, where $\Tilde{F}_{t}$ is the target corrupted flow, $i$ is the temporal interval between consecutive flows, and the length of the flow sequence is $2n+1$. The initialized flow sequence $\Tilde{F}$ are fed to the LAFC to complete the target flow $\Tilde{F}_{t}$. We denote the input of $m$-th P3D block as $\Tilde{f}^m$, and the output as $\Tilde{f}^{m+1}$. The local feature aggregation process can be formulated as.
\begin{equation}
    \Tilde{f}^{m+1} = \mbox{TC}(\mbox{SC}(\Tilde{f}^{m})) + \Tilde{f}^{m}
    \label{inertia_quant}
\end{equation} 
Where $\mbox{TC}$ represents 1D temporal convolution, and $\mbox{SC}$ is the 2D spatial convolution. We keep the temporal resolution unchanged except the final P3D block in the encoder and the P3D blocks inserted in the skip connection. In these blocks, we shrink the temporal resolution of the flow sequence to obtain the aggregated flow features of the target flow. Finally, a 2D decoder is utilized to obtain the completed target optical flow $\hat{F}_t$.

\noindent\textbf{Edge loss.} In general, flow fields are piece-wise smooth, which means the flow gradients are considerable small except motion boundaries \cite{Gao-ECCV-FGVC}. The edges in flow maps inherently contain crucial salient features that may benefit the reconstruction of object boundaries. Nevertheless, the flow completion in edge regions is a tough task, as there is no specified guidance to edge recovery. Therefore, we design a novel edge loss in LAFC to supervise the completion quality in edge regions of $\hat{F}_t$ explicitly, which can improve the flow completion quality without introducing additional computation overhead during inference.

For the completed target flow $\hat{F}_t$, we extract the edges with a small projection network $P_{e}$ and calculate the binary cross entropy loss with the edges that extracted from the ground truth $F_t$ with Canny edge detector \cite{4767851}.
\begin{equation}
    L_{e}=\mbox{BCE}(\mbox{Canny}(F_{t}), P_{e}(\hat{F}_{t}))
    \label{edge_loss}
\end{equation} 
where $L_{e}$ is the edge loss. We utilize four convolution layers with residual connection \cite{he2016deep} to formulate $P_{e}$.

\noindent\textbf{Loss function.} We adopt L1 loss to penalize $\hat{F}_{t}$ in the corrupted and the valid regions, respectively. To improve the smoothness of $\hat{F}_{t}$, we impose first and second order smoothness loss to $\hat{F}_{t}$.

What's more, we also warp the corresponding ground truth frames with $\hat{F}_{t}$ to  penalize the regions with large warp error. We adopt the L1 loss to supervise the warping quality, and expel the occlusion regions with forward-backward consistency check of ground truth optical flows for more accurate loss calculation. The loss function of LAFC is the combination of the loss terms discussed above, and the detailed formulas are provided in the supplementary material.

\subsection{Flow-guided transformer for video inpainting}
After flow completion, we propagate the content from valid regions to corrupted regions throughout the whole video to fill-in the corrupted regions that can be connected with the valid regions. The rest corrupted regions are filled with our designed flow-guided transformer (FGT). FGT takes multiple corrupted frames into consideration and synthesize these frames simultaneously. Since the motion discrepancy of completed optical flows to some extent reveals the shape and location of foreground objects and background, we integrate such information to FGT to indicate the relevant regions inside a single frame. Due to the degraded correlation between distant optical flows, the traditional all-pair interaction between tokens from distant frames may not be suitable for flow guidance integration. Therefore, we decouple MHSA along the temporal and spatial dimension, and we only integrate the flow content to the spatial MHSA. 

In both spatial and temporal transformer blocks, we introduce specific designs for efficiency and performance balance. In temporal transformer, we adopt large window to compensate the reference offset between distant frames. In spatial transformer, we divide each token map into small window based on the local smoothness prior of natural images, and supply the key and value with the condensed global tokens to perform spatial MHSA in dual perspective from local and global views.

As shown in Fig.~\ref{fig:pipeline}, given the frame sequence $\hat{X}$ after flow-guided content propagation, we crop $\hat{X}$ and completed flows into patches and project them to frame tokens $TI$ with an encoder. The completed flows are also projected to flow tokens $TF$. We refer such process as ``patch embedding".  We design interleaved temporal and spatial transformer blocks to process $TI$, and enrich the frame tokens with $TF$ in each spatial transformer block before dual perspective spatial MHSA. As for positional embedding, we follow CVPT \cite{chu2021conditional} to adopt depth-wise convolution \cite{howard2017mobilenets} after the first transformer block for video inpainting in flexible resolutions, while the pre-defined trainable positional embedding of previous works \cite{Liu_2021_FuseFormer} can only process the videos at certain resolution.

\begin{figure*}[t]
\begin{center}
\includegraphics[width=0.6\linewidth]{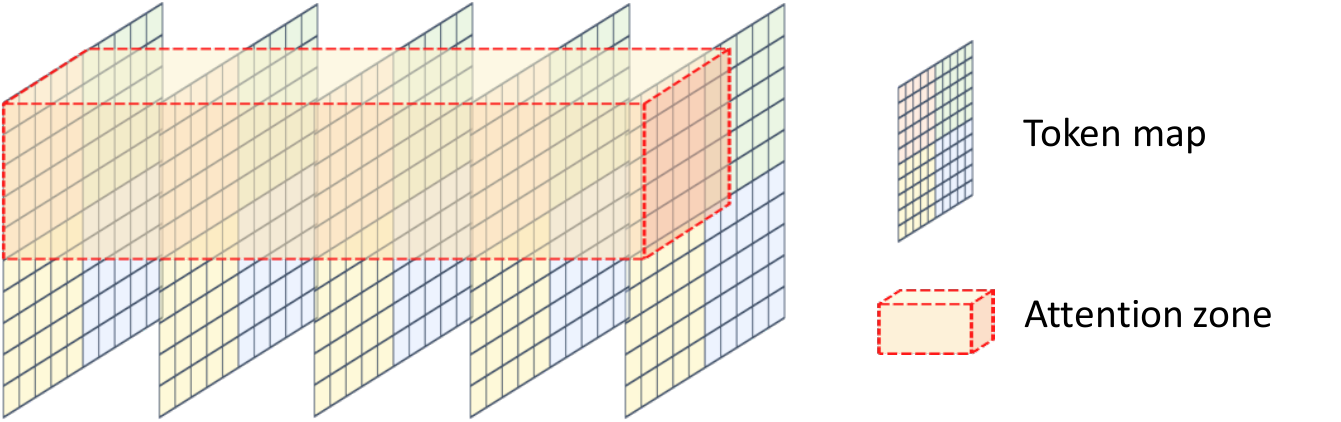}
\end{center}
   \caption{The temporal MHSA in the temporal transformer. We split non-overlapped large windows (zones) for each token map, and perform MHSA inside the cube formed by the corresponding position in each token map. The windows are shown with different colors. In this figure, we illustrate the 2$\times$2 zone.}
\label{fig:temporal}
\end{figure*}

\noindent\textbf{Temporal transformer.} In temporal transformer, attention retrieval is performed to the tokens across different frames. Since the content shifts along temporal dimension, it is reasonable to apply large size window to compensate the reference offset. Liu \etal \cite{liu2021decoupled} also demonstrate the all-pair attention strategy is unnecessary in video inpainting. Therefore, we divide each token maps in $TI$ into non-overlap cubes with large window size (denoted as ``zone") along height and width dimension and perform MHSA within the cubes, as shown in Fig.~\ref{fig:temporal}.

\begin{figure*}[t]
\begin{center}
\includegraphics[width=1\linewidth]{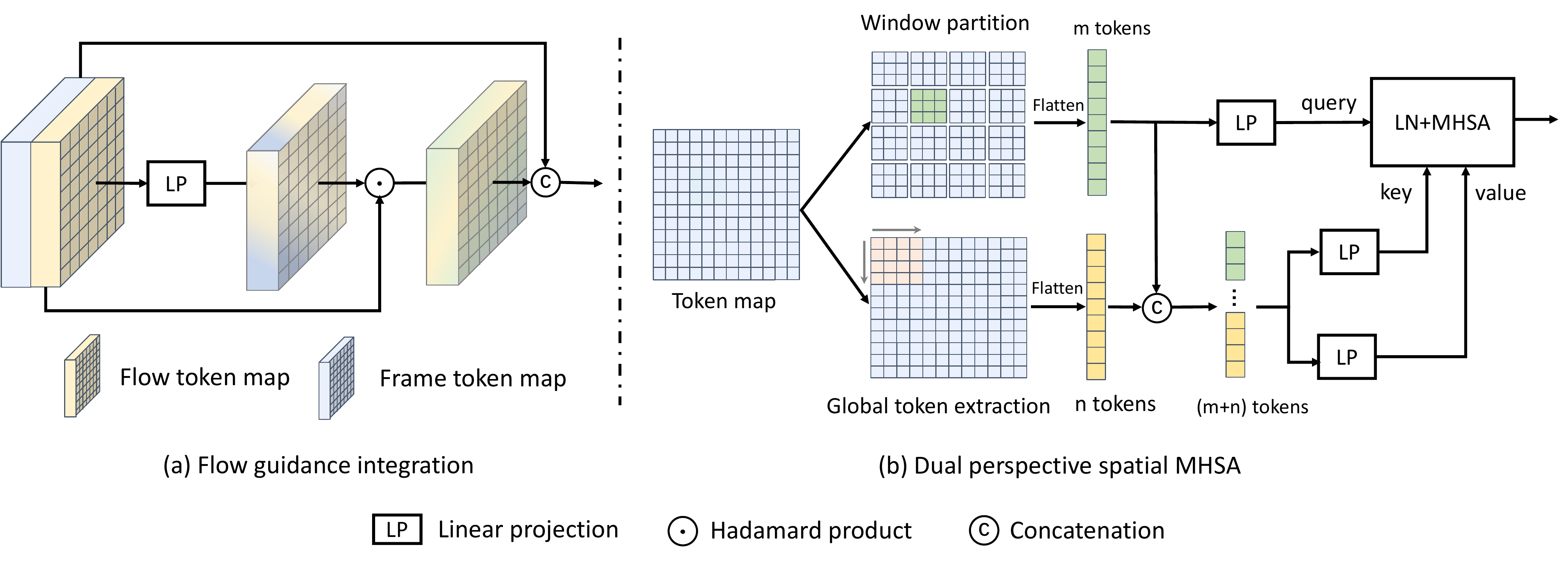}
\end{center}
   \caption{Illustration of flow guidance integration and dual perspective spatial MHSA in the spatial transformer.}
\label{fig:spatial}
\end{figure*}

\noindent\textbf{Flow guidance integration.} The motion discrepancy between different objects and background exposed by optical flows indicates the content relationship. The tokens with similar motion magnitude are more likely to be relevant. Therefore, we utilize the completed optical flows to guide the attention process in FGT.

As discussed in Sec.~\ref{p3d-unet}, optical flows are locally correlated. Therefore, we only exploit the optical flows in the spatial transformer. A straightforward way is to concatenate $TI$ and $TF$ along channel dimension directly before spatial MHSA. However, there are two problems in this way. First, the completed flows are not perfect. The distorted flows may mislead the judgement about relevant regions. Second, the appearance patches may vary a lot within objects, while the corresponding motion patterns may still be similar, which is likely to confuse the attention retrieval process. In order to ease these problems, we propose a flow-reweight module to control the impact of flow tokens $TF$ with respect to the interaction between $TF$ and $TI$, as shown in Fig.~\ref{fig:spatial}(a). We formulate the flow-reweight module as.
\begin{equation}
    \hat{TF}_t^j = TF_t^j \odot \mbox{MLP}(\mbox{Concat}(TI_t^j, TF_t^j))
\end{equation}
where $\mbox{Concat}$ is the concatenation operation. $\mbox{MLP}$ represents the MLP layers, and $\hat{TF}_t^j$ represents the $t$-th reweighted flow token map in $j$-th spatial transformer. Finally, we concatenate $\hat{TF}_t^j$ and $TI_t^j$ to obtain the flow-enhanced tokens $TK_t^j$ to enhance spatial MHSA.

\noindent\textbf{Dual perspective spatial MHSA.} We introduce window partition to spatial MHSA for efficiency. According to the local smoothness prior of natural images, the tokens are more correlated to their neighbors. Hence, we adopt relative small window size in spatial transformer. Given the $t$-th frame token map processed by $j$-th transformer $TK_t^j \in \mathbb{R}^{H \times W \times C}$, where $H$, $W$ and $C$ represent the height, width and channel size. Window partition divides $TK_t^j$ into several $h \times w$ non-overlapped windows, and MHSA is performed inside each window, respectively. However, if the window contains numerous tokens projected from corrupted regions, the attention accuracy would be deteriorated due to the lack of valid content. Therefore, we integrate global tokens to spatial MHSA. We adopt depth-wise convolution \cite{howard2017mobilenets} to condense $TK_t^j$ to global tokens, and supply them to each window. Given the kernel size $k$ and downsampling rate $s$ (also known as stride), the global tokens are generated as.
\begin{equation}
    TG_t^j=\mbox{DC}(TK_t^j, k, s)
    \label{TG_gen}
\end{equation}
where $TG_t^j$ represents the extracted global tokens and $\mbox{DC}$ is the depth-wise convolution. The query $Q_t^j(d)$, key $K_t^j(d)$ and value $V_t^j(d)$ of the $d$-th window in $TK_t^j$ are generated as.
\begin{equation}
\begin{aligned}
    & Q_t^j(d) = \mbox{MLP}(\mbox{LN}(TK_t^j(d))) \\
    & K_t^j(d) = \mbox{MLP}(\mbox{LN}(\mbox{Concat}(TK_t^j(d),TG_t^j))) \\
    & V_t^j(d) = \mbox{MLP}(\mbox{LN}(\mbox{Concat}(TK_t^j(d),TG_t^j)))
\end{aligned}
\end{equation}
where $TK_t^j(d)$ represents the $d$-th window in $TK_t^j$, and $\mbox{LN}$ is layer normalization \cite{ba2016layer}. After we obtain $Q_t^j(d)$, $K_t^j(d)$ and $V_t^j(d)$, we apply spatial MHSA to process them. The dual perspective spatial MHSA is illustrated in Fig.~\ref{fig:spatial}(b).

Note that the global tokens are shared by all the windows. In each spatial transformer, if we adopt all-pair attention retrieval for MHSA, each token will retrieve $H \times W$ tokens. While the token number for retrieval in our dual perspective spatial MHSA is $(\lceil\frac{H}{s}\rceil \times \lceil\frac{W}{s}\rceil + h \times w)$. It is easy to derive that when $s \geq \lceil\sqrt{\frac{HW}{HW-hw}}\rceil$, the referenced token number will be smaller than the token number in all-pair attention retrieval.

Recently, focal transformer \cite{yang2021focal} has also adopted the combination of local and global attention in transformer. Compared with \cite{yang2021focal}, our method decouples the global token size and the window shape, which is more flexible than the sub-window pooling strategy in focal transformer.

\noindent\textbf{Loss function} We adopt the reconstruction loss in the corrupted and the valid regions together with the T-Patch GAN loss \cite{chang2019free} to supervise the training process. We use hinge loss as the adversarial loss. We provide the detailed loss formulas in the supplementary material.


\section{Experiments}

\subsection{Settings}
We adopt Youtube-VOS \cite{xu2018youtube} and DAVIS \cite{caelles20182018} datasets for evaluation. Youtube-VOS contains 4453 videos and DAVIS contains 150 videos. We adopt the training set of Youtube-VOS to train our networks. As for Youtube-VOS, we evaluate the trained models on its testset. Since DAVIS contains densely annotated masks on its training set, we adopt its training set to evaluate our method.

Following the previous work \cite{Gao-ECCV-FGVC}, we choose PSNR, SSIM \cite{wang2004image} and LPIPS \cite{zhang2018unreasonable} as our evaluation metrics. Meanwhile, we adopt end-point-error (EPE) to evaluate the flow completion quality. We compare our method with state-of-the-art baselines, including VINet \cite{kim2019deep}, DFGVI \cite{Xu_2019_CVPR}, CPN \cite{lee2019cpnet}, OPN \cite{Oh_2019_ICCV}, 3DGC \cite{chang2019free}, STTN \cite{yan2020sttn}, FGVC \cite{Gao-ECCV-FGVC}, TSAM \cite{zou2020progressive}, DSTT \cite{liu2021decoupled} and FFM \cite{Liu_2021_FuseFormer}.  

\subsection{Implementation details}
In our experiments, We utilize RAFT \cite{teed2020raft} to extract optical flows. In flow completion network, the flow interval and input flow number are both set to 3. The flow locating in middle of the local temporal window is treated as the target flow for completion. We adopt gradient propagation \cite{Gao-ECCV-FGVC} as our flow-guided content propagation strategy, and the detailed procedure will be provided in the supplementary material. As for FGT, we keep the patch embedding method the same as FFM \cite{Liu_2021_FuseFormer} for fair comparisons. We utilize the forward optical flows in the flow guidance integration module. FGT adopts 8 transformer blocks in total (4 temporal and 4 spatial transformer blocks). In the temporal transformer, we adopt 2$\times$2 zone division for temporal MHSA. In the spatial transformer, the downsampling rate of the global token is 4, while the window size is 8. We adopt Adam optimizer \cite{kingma2014adam} to train our networks. The training iteration is 280K for LAFC and 500K for FGT. The initial learning rate is 1$e$-4, which is divided by 10 after 120K iterations for LAFC and 300K iterations for FGT. For ablation studies, following FFM \cite{Liu_2021_FuseFormer}, we train FGT for 250K iterations, and the learning rate is divided by 10 after 200K iterations. We perform ablation studies on DAVIS dataset.

\begin{table*}[t]
\begin{center}
\scriptsize
\caption{Quantitative results on the Youtube-VOS and DAVIS datasets. The best and second best numbers for each metric are indicated by \textcolor[rgb]{1.00,0.00,0.00}{\underline{red}} and \textcolor[rgb]{0.00,0.00,1.00}{\underline{blue}} fonts, respectively. $\downarrow$ means lower is better, while $\uparrow$ means higher is better. ``FGT" represents we adopt our proposed flow-guided transformer to fill all the pixels in the corrupted regions without flow-guided content propagation.}
\label{quan_rets}
\begin{tabular}{@{}lccccccccc@{}}
\toprule
\multirow{3}{*}{Method} & \multicolumn{3}{c}{\multirow{2}{*}{Youtube-VOS}} & \multicolumn{6}{c}{DAVIS} \\ \cline{5-10}
& \multicolumn{3}{c}{} & \multicolumn{3}{c}{square} & \multicolumn{3}{c}{object} \\ \cmidrule(l){2-4} \cmidrule(l){5-7} \cmidrule(l){8-10}
& PSNR$\uparrow$ & SSIM$\uparrow$ & LPIPS$\downarrow$ & PSNR$\uparrow$ & SSIM$\uparrow$ & LPIPS$\downarrow$ & PSNR$\uparrow$ & SSIM$\uparrow$ & LPIPS$\downarrow$ \\ \midrule
VINet \cite{kim2019deep}& 29.83 & 0.955 & 0.047 & 28.32 & 0.943 & 0.049 & 28.47 & 0.922 & 0.083 \\
DFGVI \cite{Xu_2019_CVPR} &32.05 & 0.965 & 0.038 & 29.75 & 0.959 & 0.037 & 30.28 & 0.925 & 0.052 \\
CPN \cite{lee2019cpnet} &32.17 & 0.963 & 0.040 & 30.20 & 0.953 & 0.049 & 31.59 & 0.933 & 0.058 \\
OPN \cite{Oh_2019_ICCV} &32.66 & 0.965 & 0.039 & 31.15 & 0.958 & 0.044 & 32.40 & 0.944 & 0.041 \\
3DGC \cite{chang2019free} &30.22 & 0.961 & 0.041 & 28.19 & 0.944 & 0.049 & 31.69 & 0.940 & 0.054 \\
STTN \cite{yan2020sttn} &32.49 & 0.964 & 0.040 & 30.54 & 0.954 & 0.047 & 32.83 & 0.943 & 0.052  \\
TSAM \cite{zou2020progressive}& 31.62 & 0.962 & 0.031 & 29.73 & 0.951 & 0.036 & 31.50 & 0.934 & 0.048 \\
DSTT \cite{liu2021decoupled}& 33.53 & 0.969 & 0.031 & 31.61 & 0.960 & 0.037 & 33.39 & 0.945 & 0.050 \\
FFM \cite{Liu_2021_FuseFormer}& 33.73 & 0.970 & 0.030 & 31.87 & 0.965 & 0.034 & 34.19 & 0.951 & 0.045 \\
FGT & \textcolor[rgb]{0.00,0.00,1.00}{\underline{34.04}} & 0.971 & 0.028 & \textcolor[rgb]{0.00,0.00,1.00}{\underline{32.60}} & 0.965 & 0.032 & \textcolor[rgb]{0.00,0.00,1.00}{\underline{34.30}} & 0.953 & 0.040 \\
FGVC \cite{Gao-ECCV-FGVC} & 33.94 & \textcolor[rgb]{0.00,0.00,1.00}{\underline{0.972}} & \textcolor[rgb]{0.00,0.00,1.00}{\underline{0.026}} & 32.14 & \textcolor[rgb]{0.00,0.00,1.00}{\underline{0.967}} & \textcolor[rgb]{0.00,0.00,1.00}{\underline{0.030}} & 33.91 & \textcolor[rgb]{0.00,0.00,1.00}{\underline{0.955}} & \textcolor[rgb]{0.00,0.00,1.00}{\underline{0.036}} \\
Ours &\textcolor[rgb]{1.00,0.00,0.00}{\underline{34.53}} & \textcolor[rgb]{1.00,0.00,0.00}{\underline{0.976}} & \textcolor[rgb]{1.00,0.00,0.00}{\underline{0.024}} & \textcolor[rgb]{1.00,0.00,0.00}{\underline{33.41}}& \textcolor[rgb]{1.00,0.00,0.00}{\underline{0.974}} & \textcolor[rgb]{1.00,0.00,0.00}{\underline{0.023}} & \textcolor[rgb]{1.00, 0.00, 0.00}{\underline{34.96}} & \textcolor[rgb]{1.00,0.00,0.00}{\underline{0.966}} & \textcolor[rgb]{1.00,0.00,0.00}{\underline{0.029}}  \\
\bottomrule
\end{tabular}
\end{center}
\end{table*}

\subsection{Quantitative evaluation}
During inference, the resolution of videos is set to 432$\times$256. We generate square masksets with continuous motion trace for Youtube-VOS and DAVIS datasets. The average size of the masks in the square maskset is $\frac{1}{16}$ of the whole frame. We shuffle DAVIS object maskset randomly and corrupt frames with these masks to evaluate video inpainting performance upon object masks. For fair comparisons among flow-based video inpainting methods, we utilize the same optical flow extractor for DFGVI \cite{Xu_2019_CVPR}, FGVC \cite{Gao-ECCV-FGVC} and our method. 

We report the quantitative evaluation results of our method and other baselines in Tab.~\ref{quan_rets}. Our method outperforms previous baselines by a significant margin on all three metrics, which means the restored videos from our method enjoy less distortion and better perceptual quality than previous counterparts. What's more, if we fill the corrupted region purely with FGT, we can still outperform previous transformer-based video inpainting baselines \cite{yan2020sttn,Liu_2021_FuseFormer,liu2021decoupled}.

\begin{figure*}[t]
\captionsetup{font={scriptsize}}
    \centering
    \subcaptionbox{Input\label{subfig:input}}[0.129\linewidth]
    {
        \includegraphics[width=1\linewidth]{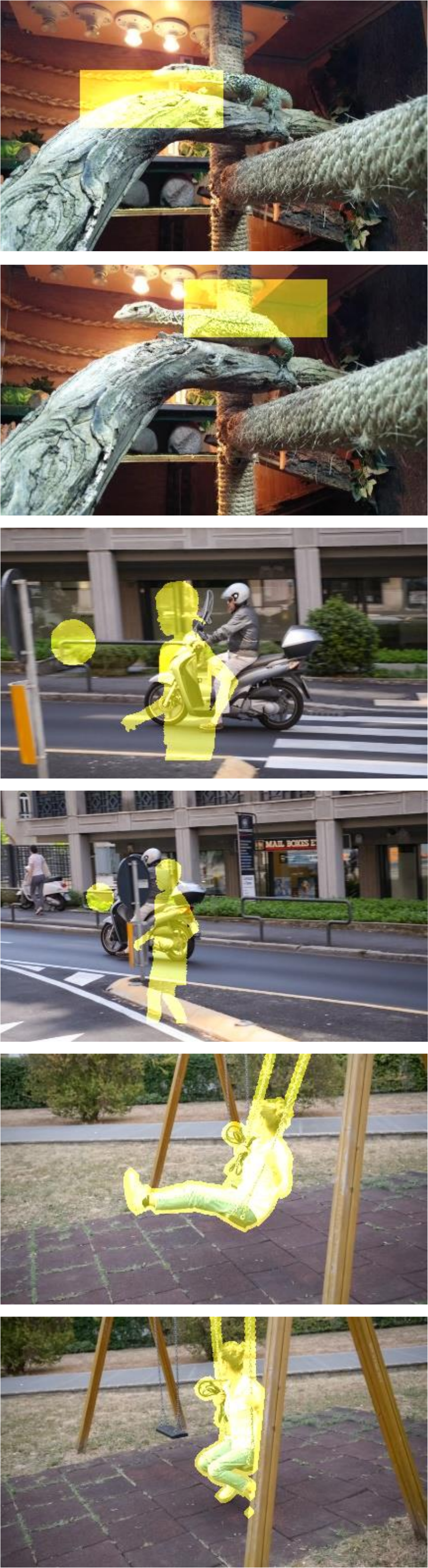}
    }
    \subcaptionbox{STTN \cite{yan2020sttn}\label{subfig:sttn}}[0.129\linewidth]
    {
        \includegraphics[width=1\linewidth]{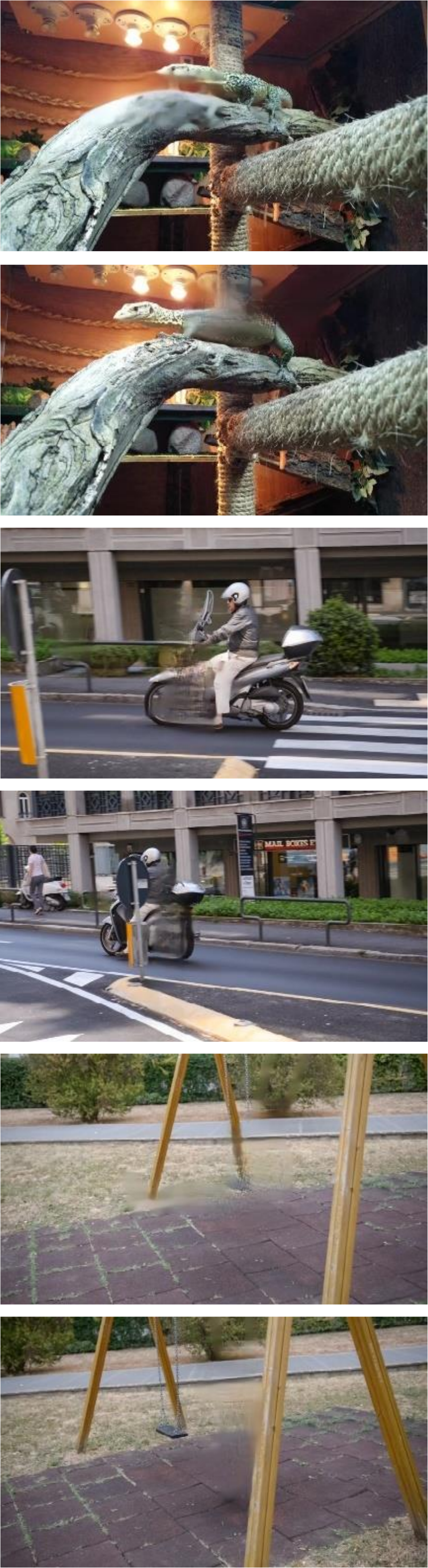}
    }
    \subcaptionbox{TSAM \cite{zou2020progressive} \label{subfig:tsam}}[0.129\linewidth]
    {
        \includegraphics[width=1\linewidth]{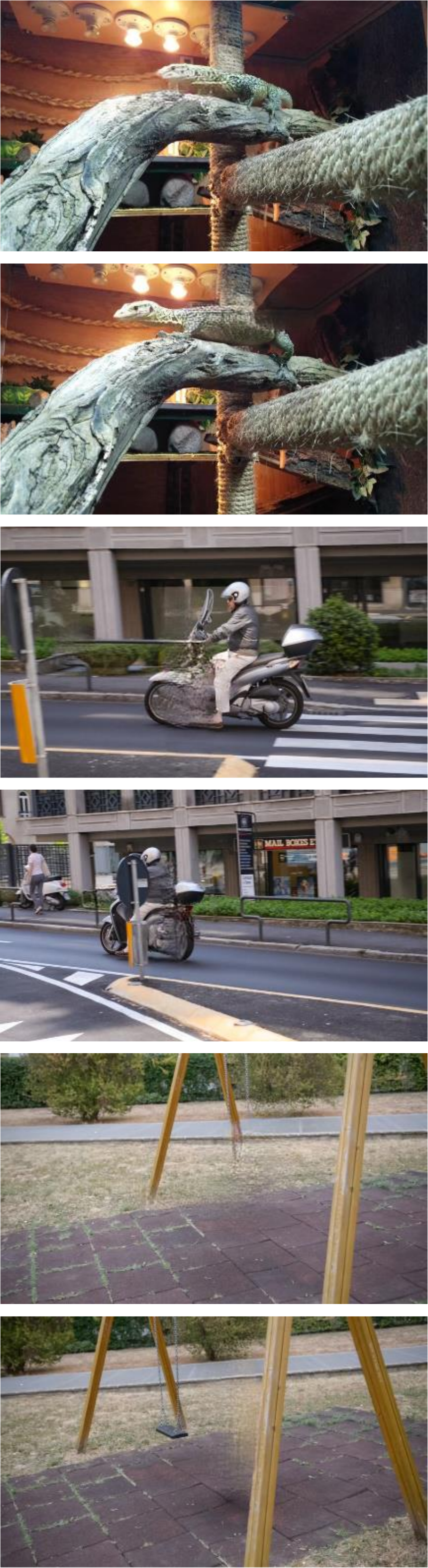}
    }
    \subcaptionbox{DSTT \cite{liu2021decoupled} \label{subfig:dstt}}[0.129\linewidth]
    {
        \includegraphics[width=1\linewidth]{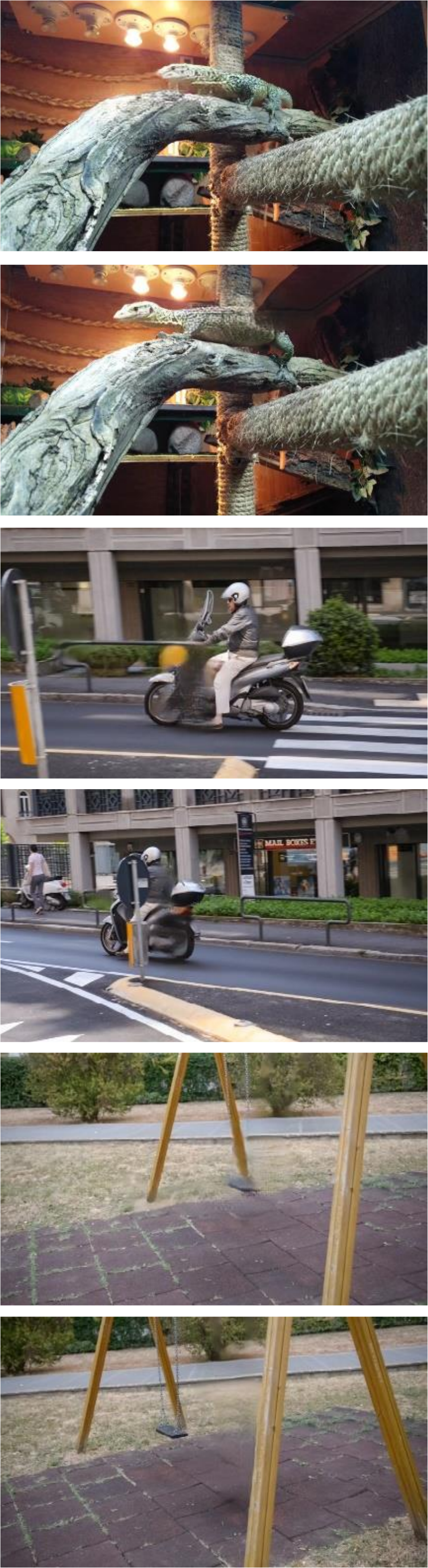}
    }
    \subcaptionbox{FFM \cite{Liu_2021_FuseFormer} \label{subfig:ffm}}[0.129\linewidth]
    {
        \includegraphics[width=1\linewidth]{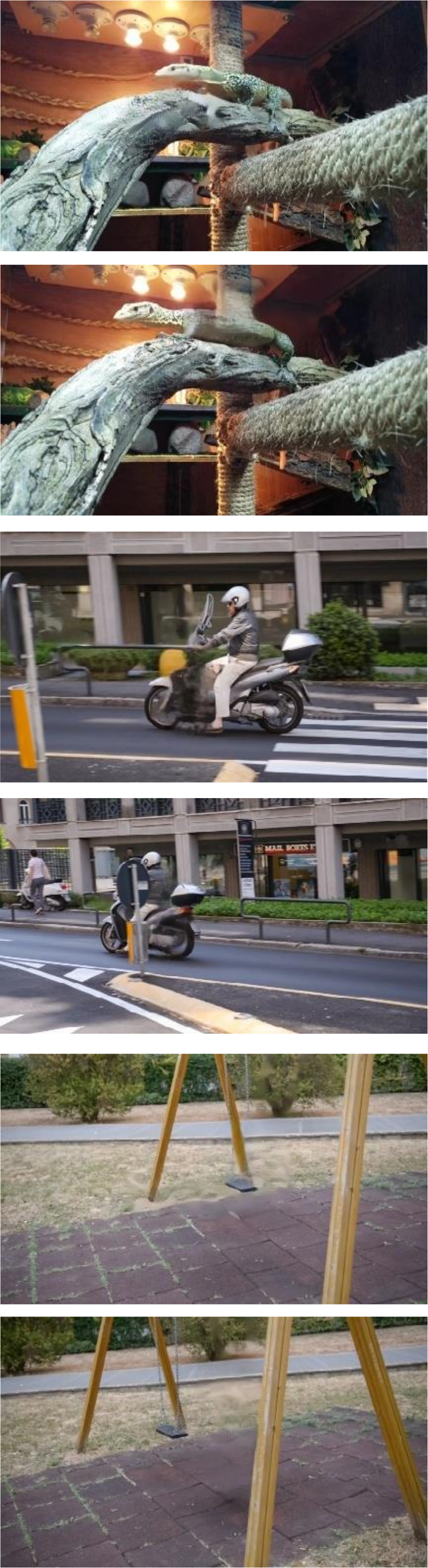}
    }
    \subcaptionbox{FGVC \cite{Gao-ECCV-FGVC} \label{subfig:fgvc}}[0.129\linewidth]
    {
        \includegraphics[width=1\linewidth]{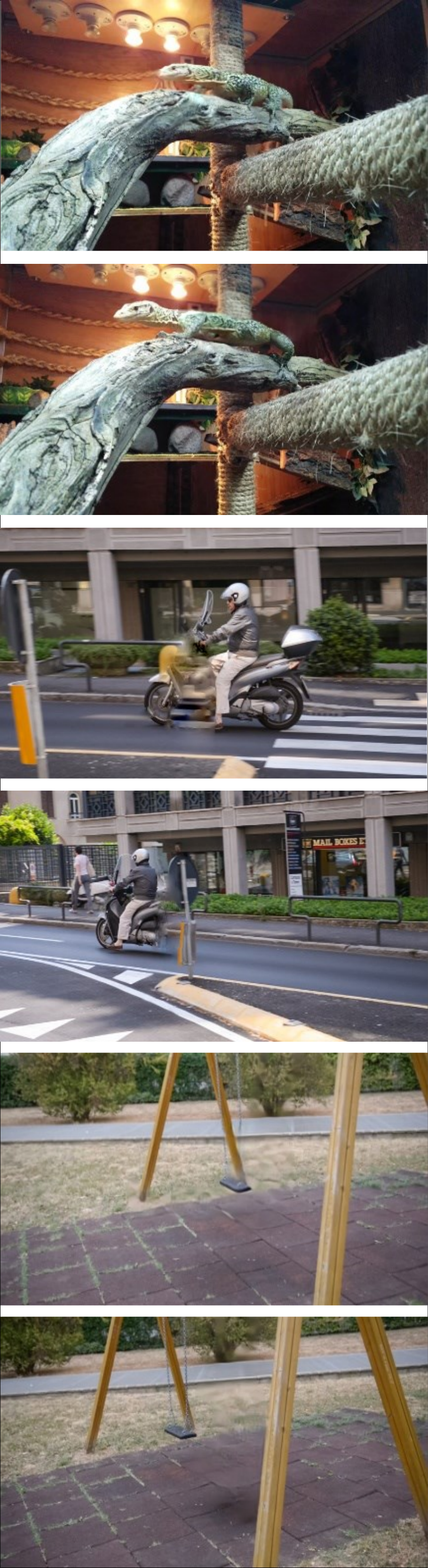}
    }
    \subcaptionbox{Ours \label{subfig:ours}}[0.129\linewidth]
    {
        \includegraphics[width=1\linewidth]{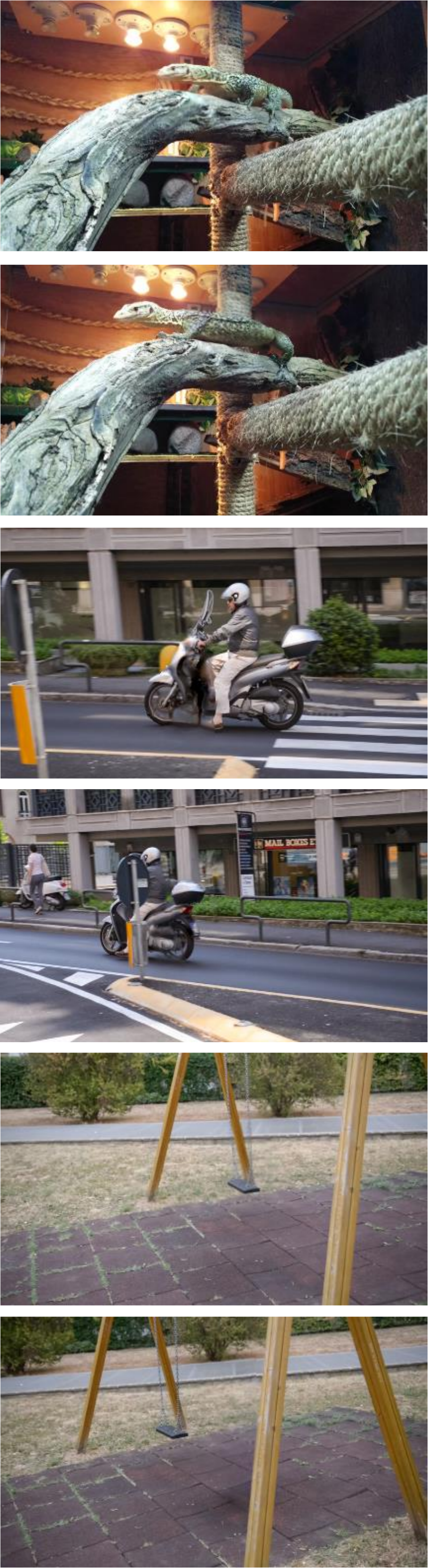}
    }
    \captionsetup{font={normalsize}}
    \caption{Qualitative comparison between our method and some recent baselines \cite{yan2020sttn,Gao-ECCV-FGVC,zou2020progressive,liu2021decoupled,Liu_2021_FuseFormer}. From top to bottom, every two rows display inpainting results of square mask set, object mask set, and object removal, respectively.}
    \label{quali_rets}
\end{figure*}

\subsection{Qualitative comparisons}
We compare the qualitative results between our method and five recent baselines \cite{yan2020sttn,Gao-ECCV-FGVC,zou2020progressive,liu2021decoupled,Liu_2021_FuseFormer} under the square mask, object mask and object removal settings. The results are shown in Fig.~\ref{quali_rets}. Compared with these baselines, our method enjoys outstanding visual quality. Our method can complete more accurate optical flows, which describes the motion trajectory with high fidelity. Therefore, our method enjoys less distortion in the content propagation stage than FGVC \cite{Gao-ECCV-FGVC}. What's more, the completed optical flows provide accurate object clusters. Such information leads to more accurate attention retrieval and naturally produce better visual quality. We will provide more video inpainting results in the supplementary materials.

\begin{figure*}[t]
\begin{center}
\includegraphics[width=1\linewidth]{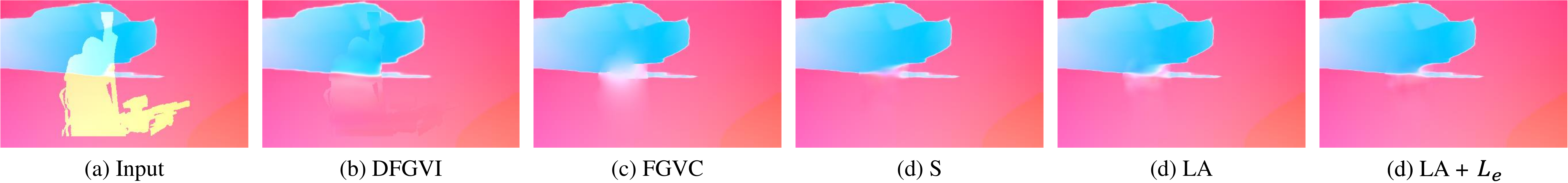}
\end{center}
   \caption{Comparison of flow results between DFGVI \cite{Xu_2019_CVPR}, FGVC \cite{Gao-ECCV-FGVC}, and several variants of our method. S: single flow completion, LA: Flow completion with local aggregation, $L_e$: Edge loss.}
\label{fig:flowCompare}
\end{figure*}

\begin{figure*}[t]
\begin{center}
\includegraphics[width=1\linewidth]{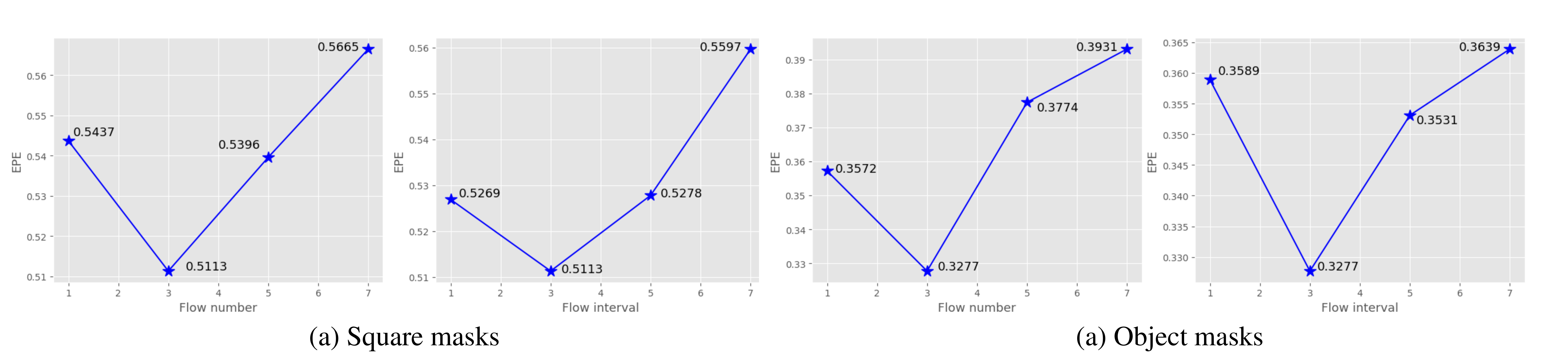}
\end{center}
   \caption{EPE results with varying flow number (when flow interval is 3) or varying flow interval (when flow number is 3) on both square and object mask sets.}
\label{LAFC}
\end{figure*}

\begin{table}[t]
    \caption{
    \textbf{Model analysis}
    We report the analysis of the method variants and the comparison of the efficiency between FGT and other baselines.
    }
    \label{tab:ablation}
    \centering
    
    \begin{minipage}{0.53\linewidth}
    \centering
    {(a) \textbf{Analysis about method variants.}
    } 
    \end{minipage}
    \hfill
    \begin{minipage}{0.35\linewidth}
    {(b) \textbf{Efficiency analysis.} 
    }
    \end{minipage}
    \hfill
    \mpage{0.53}{
    \resizebox{1\linewidth}{!} 
    {
    \begin{tabular}{@{}ccccccc@{}}
    \toprule
    \multirow{2}{*}{Method} & \multicolumn{3}{c}{square} & \multicolumn{3}{c}{object} \\ \cmidrule(r){2-4} \cmidrule(r){5-7}
    & PSNR$\uparrow$ & SSIM$\uparrow$ & LPIPS$\downarrow$ & PSNR$\uparrow$ & SSIM$\uparrow$ & LPIPS$\downarrow$ \\ \midrule
    FGVC \cite{Gao-ECCV-FGVC} & 32.14 & 0.967 & 0.030 & 33.91 & 0.955 & 0.034 \\
    FGVC+FGT & 32.49 & 0.968 & 0.027 & 34.58 & 0.956 & 0.031 \\
    LAFC+FGT & \textbf{33.41} & \textbf{0.974} & \textbf{0.023} & \textbf{34.96} & \textbf{0.966} & \textbf{0.029} \\
    \bottomrule
    \end{tabular}
    }
}
\hfill
\mpage{0.4}{
\centering
    \resizebox{1\linewidth}{!} 
    {
    \begin{tabular}{@{}lccc@{}}
    \toprule
    Method & FLOPs(per frame) & Params & Speed \\ \midrule
    STTN \cite{yan2020sttn} & 477.91G      & 16.56M       & 0.22s        \\
    FFM \cite{Liu_2021_FuseFormer} &  579.82G      & 36.59M       & 0.30s        \\
    FGT(all-pair) & 703.22G & 42.31M & - \\
    FGT & 455.91G & 42.31M & 0.39s \\      \bottomrule   
    \end{tabular}
    }
}
\label{Tab:model_analysis}
\end{table}

\subsection{Ablation Studies}
\noindent \textbf{Model analysis.} We compare our method with (1) FGVC and (2) FGVC+ \\FGT to justify the design of our method over flow completion and image inpainting baseline \cite{yu2018generative}. The results in Tab.~\ref{Tab:model_analysis}(a) demonstrate the effectiveness of our method in both flow completion and frame synthesis. In Tab.~\ref{Tab:model_analysis}(b), we compare FGT with different transformer baselines. Since FLOPs in video inpainting is related to the number of frames processed simultaneously, we assume the processed frame number is 20, which is a common practice in STTN \cite{yan2020sttn} and FFM \cite{Liu_2021_FuseFormer}.``FGT(all-pair)" means we adopt all-pair attention in FGT, which consumes much more computation overhead compared with FGT. If we adopt flow-guided content propagation, we can obtain better video inpainting quality, but the speed will degrade to 2.11s/frame, which indicates the performance-efficiency trade-off in our method. We provide detailed run-time analysis in the supplementary material.

\noindent \textbf{Local flow aggregation and edge loss for flow completion.} We report the end-point-error (EPE) of single flow completion (replace the P3D blocks with vanilla convolution blocks), local aggregation for flow completion without and with edge loss, together with two baselines \cite{Xu_2019_CVPR,Gao-ECCV-FGVC} in Tab.~\ref{flow_comp2}. With the introduction of local aggregation and edge loss, our method achieves substantial improvement. The subjective results are shown in Fig.~\ref{fig:flowCompare}. With local aggregation, our method can exploit the complementary flow features in a local temporal window, which is beneficial to complete accurate flow shape. With edge loss, our method can synthesize optical flows with clearer motion boundaries. Finally, we report the influence of flow number and flow interval w.r.t. EPE in Fig.~\ref{LAFC}. When the flow number or interval is too small, the target flow cannot utilize abundant references for accurate flow completion, which undermines the performance. When the flow number or interval is large, the flow completion performance deteriorates gradually, which reveals the distant flows contribute less to flow completion relative to local flows.

\begin{figure*}[t]
\begin{center}
\includegraphics[width=1\linewidth]{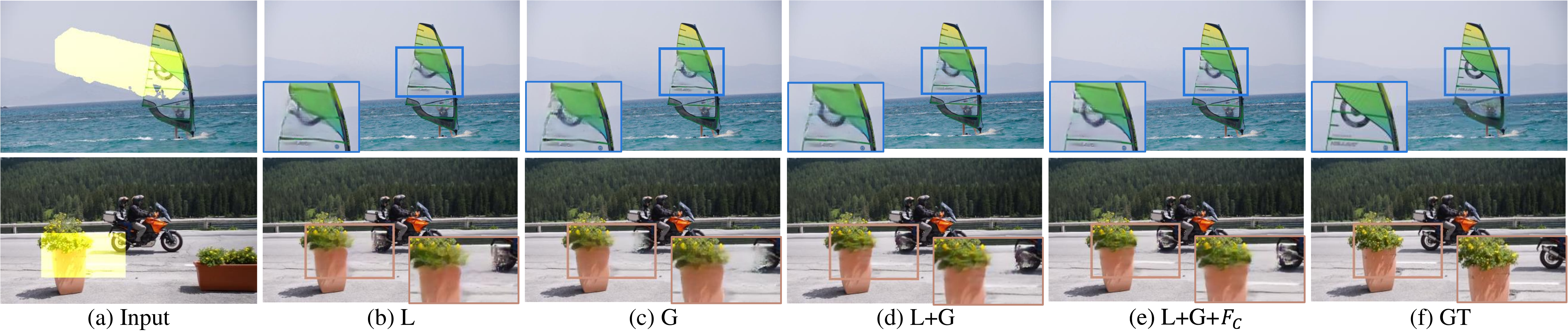}
\end{center}
   \caption{Qualitative comparison of different components in the dual perspective transformer. L: Local window attention. G: Global tokens. $F_{c}$: Flow guidance with flow-reweight module.}
\label{fig:transCompare}
\end{figure*}

\begin{table}[t]
\fontsize{8}{9}\selectfont
\begin{center}
\caption{Ablation study about flow completion. S: single flow completion, LA: Flow completion with local aggregation, $L_e$: Edge loss.}
\label{flow_comp2}
\setlength{\tabcolsep}{0.4mm}{
\begin{tabular}{@{}lccccc@{}}
\toprule
\multirow{2}{*}{Maskset} & \multicolumn{5}{c}{EPE$\downarrow$} \\ \cmidrule(r){2-6}
& DFGVI \cite{Xu_2019_CVPR} & FGVC \cite{Gao-ECCV-FGVC} & S & LA & LA + $L_e$ \\ \midrule
square & 1.161 & 0.633 & 0.546 & 0.524 & 0.511 \\
object & 1.053 & 0.491 & 0.359 & 0.338 & 0.328 \\
\bottomrule
\end{tabular}}
\end{center}
\end{table}

\begin{figure*}[t]
\begin{center}
\includegraphics[width=1\linewidth]{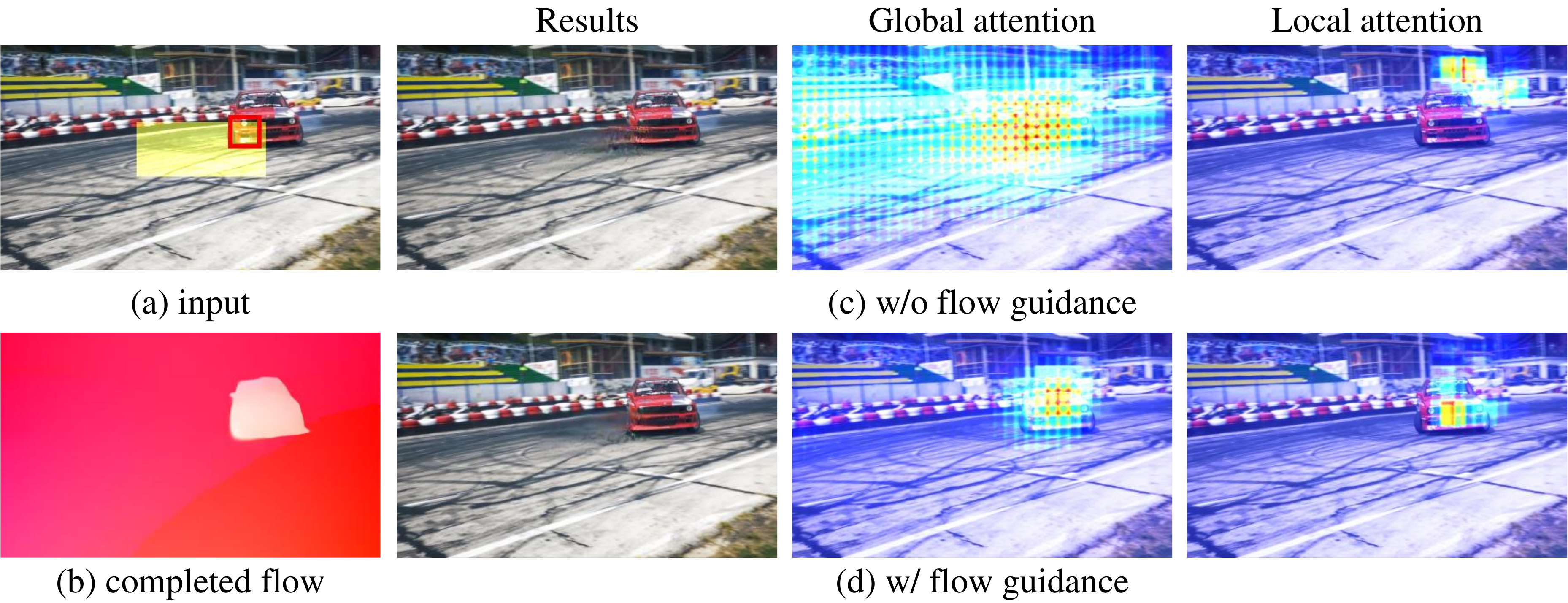}
\end{center}
   \caption{Attention map visualization of our transformer model with/without the flow guidance. The red square in (a) indicates the location of the chosen query token for visualization.}
\label{fig:att_vis}
\end{figure*}

\begin{table}[t]
\fontsize{8}{9}\selectfont
\begin{center}
\caption{Ablation study about the spatial transformer. W: Local window partition. G: Global tokens. F$_{C}$: Completed flow tokens. RF: Flow-reweight module.}
\label{fgdp_trans}
\setlength{\tabcolsep}{1mm}{
\begin{tabular}{@{}lccccccccc@{}}
\toprule
\multirow{2}{*}{W} & \multirow{2}{*}{G} & \multirow{2}{*}{F$_{C}$} & \multirow{2}{*}{RF} & \multicolumn{3}{c}{square} & \multicolumn{3}{c}{object} \\ \cmidrule(r){5-7} \cmidrule(r){8-10}
&  &  &  &  PSNR$\uparrow$ & SSIM$\uparrow$ & LPIPS$\downarrow$ & PSNR$\uparrow$ & SSIM$\uparrow$ & LPIPS$\downarrow$ \\ \midrule
\checkmark & - & - & - & 31.37 & 0.957 & 0.038 & 32.98 & 0.945 & 0.051 \\
- & \checkmark & - & - & 31.42 & 0.958 & 0.040 & 33.10 & 0.945 & 0.050 \\
\checkmark & \checkmark & - & - & 31.62 & 0.959 & 0.038 & 33.25 & 0.946 & 0.048 \\
\checkmark & \checkmark & \checkmark & - & 31.54 & 0.958 & 0.039 & 33.12 & 0.945 & 0.049 \\
\checkmark & \checkmark & \checkmark & \checkmark & 31.87 & 0.961 & 0.036 & 33.52 & 0.947 & 0.045 \\
\bottomrule
\end{tabular}}
\end{center}
\end{table}

\noindent \textbf{Flow guidance integration and dual perspective spatial MHSA.} In this part, we adopt the transformer to synthesize all the pixels in the corrupted regions for fair comparisons across different settings. We evaluate the effectiveness of the dual perspective tokens, the completed flow guidance and the flow-reweight module in spatial MHSA, and report the corresponding results in Tab.~\ref{fgdp_trans}. 

The quantitative results demonstrate the effectiveness of our proposed method. Compared with attention with only local or global tokens, the combination of these two perspective tokens achieves significant performance boost. With the introduction of the completed flow tokens and the flow-reweight module, the performance of our model boosts further. When we remove the flow-reweight module, the performance degrades, which demonstrates the necessity to introduce flow guidance and control its impact during attention retrieval. 

The qualitative comparisons between different components in our flow-guided transformer are shown in Fig.~\ref{fig:transCompare}. We can observe the substantial improved visual quality on dual perspective attention and the introduction of flow guidance. The combination of global and local tokens enlarges the attention retrieval space while maintaining the local smoothness simultaneously. As for flow guidance, we visualize the local and global attention maps in Fig.~\ref{fig:att_vis}. The red square in Fig.~\ref{fig:att_vis}(a) indicates the query token. With flow guidance, our transformer tends to query the tokens with similar motion pattern (e.g. tokens in car region), which leads to clearer object boundary for video inpainting in higher quality. 

\section{Conclusion}
In this work, we propose a flow-guided transformer for video inpainting, which introduces a novel way to leverage the motion discrepancy from optical flows to instruct the attention retrieval in transformer. We decouple the attention module along spatial and temporal dimension to facilitate the integration of the completed flows. We propose the flow-reweight module to control the impact of the flows in the attention retrieval process. What's more, in both temporal and spatial transformer blocks, we design specific window partition strategy for better efficiency while maintaining the competitive performance. Besides the proposed flow-guided transformer, We design a flow completion network to exploit the complementary features of the optical flows in a local temporal window, and introduce edge loss to supervise the reconstruction of flows for clear motion boundaries. The high-quality completed flows benefit the content propagation and flow-guided transformer. Extensive experiments demonstrate the effectiveness of our proposed method. 

\flushleft\textbf{Acknowledgment.} This work was supported by the Natural Science Foundation of China under Grants 62036005, 62022075, and 62021001, and by the Fundamental Research Funds for the Central Universities under Contract No. WK3490000006.

\clearpage
%
%
\bibliographystyle{splncs04}
\bibliography{egbib}
\end{document}